\documentclass[journal]{vgtc}                % final (journal style)
\ifpdf%                                % if we use pdflatex
  \pdfoutput=1\relax                   % create PDFs from pdfLaTeX
  \usepackage{graphicx}                % allow us to embed graphics files
  \DeclareGraphicsExtensions{.pdf,.png,.jpg,.jpeg} % for pdflatex we expect .pdf, .png, or .jpg files
\else%                                 % else we use pure latex
  \usepackage{graphicx}                % allow us to embed graphics files
  \DeclareGraphicsExtensions{.eps}     % for pure latex we expect eps files
\fi%

\graphicspath{{figures/}{pictures/}{images/}{./}} % where to search for the images

\usepackage{microtype}                 % use micro-typography (slightly more compact, better to read)
\PassOptionsToPackage{warn}{textcomp}  % to address font issues with \textrightarrow
\usepackage{textcomp}                  % use better special symbols
\usepackage{mathptmx}                  % use matching math font
\usepackage{times}                     % we use Times as the main font
\usepackage{cite}                      % needed to automatically sort the references
\usepackage{tabu}                      % only used for the table example
\usepackage{booktabs}                  % only used for the table example

\usepackage{graphicx}
\usepackage{hyperref}
\usepackage{tikz}
\usepackage{comment}
\usepackage{amsmath,amssymb} % define this before the line numbering.
\usepackage{color}
\usepackage{nicefrac}
\usepackage{multirow}
\usepackage{placeins}
\usepackage[normalem]{ulem}
\newcommand{\etal}{\textit{et al}.}
\newcommand{\ie}{\textit{i}.\textit{e}., }
\newcommand{\eg}{\textit{e}.\textit{g}. }

\newcommand{\spheading}[2][3.2em]{\raisebox{-0em}{\rotatebox{90}{\parbox{#1}{\raggedright #2}}}}

\onlineid{0}

\vgtccategory{Research}
\newcommand{\rev}[1]{ \color{black} #1 \color{black}}
\vgtcpapertype{algorithm/technique}

\title{EllSeg: An Ellipse Segmentation Framework for Robust Gaze Tracking}

\author{Rakshit S. Kothari$^{*}$, Aayush K. Chaudhary$^{*}$, Reynold J. Bailey, Jeff B. Pelz and Gabriel J. Diaz}
\authorfooter{
\item
All authors are with the Rochester Institute of Technology.
\item[*]
R.S.Kothari and A.K. Chaudhary contributed equally to this work.
}

\abstract{Ellipse fitting, an essential component in pupil or iris tracking based video oculography, is performed on previously segmented eye parts generated using various computer vision techniques. Several factors, such as occlusions due to eyelid shape, camera position or eyelashes, frequently break ellipse fitting algorithms that rely on well-defined pupil or iris edge segments. In this work, we propose training a convolutional neural network to directly segment entire elliptical structures and demonstrate that such a framework is robust to occlusions and offers superior pupil and iris tracking performance (at least 10$\%$ and 24$\%$ increase in pupil and iris center detection rate respectively within a two-pixel error margin) compared to using standard eye parts segmentation for multiple publicly available synthetic segmentation datasets.} 


\vgtcinsertpkg

\begin{document}

%% The ``\maketitle'' command must be the first command after the
%% ``\begin{document}'' command. It prepares and prints the title block.

%% the only exception to this rule is the \firstsection command
\firstsection{Introduction}

\maketitle
% The inclusion of head-mounted video-oculography in Augmented/Virtual Reality (AR/VR) applications has demonstrated advantages, from potentially mitigating Vergence-Accommodation conflict~\cite{padmanaban2017optimizing} to Infinite VR Walk~\cite{sun2018towards}}. A common approach in head-mounted video-oculography involves ...

There is great potential for the use of eye tracking in augmented and virtual reality (AR/VR) displays both as a means for user interaction, and for gaze-dependent rendering techniques that can both increase visual fidelity~\cite{cholewiak2017chromablur} while also lowering computational overhead~\cite{patney2016towards}. Contemporary methods for eye tracking in VR and AR build upon techniques established in the context of head-mounted video-oculography, which involve the use of one or more infrared light sources placed next to infrared \textit{eye} cameras.  These eye cameras are pointed towards each of the wearer's eyes while a third camera, referred to as the scene camera, points away from the wearer to capture the environment being observed~\cite{Duchowski2007a}. Existing solutions extract gaze descriptive features such as pupil center~\cite{Fuhl2015a,Fuhl2016e,Fuhl2017c,Santini2018b,Santini2018c,Kim2019b}, pupil ellipse~\cite{Fuhl2017d,Swirski2012a,Swirski2013d,Li2018,Yiu2019b}, iris ellipse~\cite{Wood2014b,Wu2004,Plopski2015}, or track iridial features~\cite{Chaudhary2019c,Ong2010}. These solutions vary in algorithmic complexity, latency, and computational power requirements. Extracted features are then correlated to a measure of gaze using calibration routines~\cite{itoh2014interaction,binaee2016binocular,nystrom2013influence}, which compensate for person-specific physiological differences.

% Gabe started writing this, and then decided it was not necessary.
% Although many current methods for eye tracking in VR/AR follow these principles closely, integrations into VR displays are unique in that they do not map gaze onto imagery recorded from a scene camera. Instead, the user or provided with a digital record of gaze direction within head (and possibly world) centered coordinates to be used for additional computation and algorithmic investigation.  }

Despite many recent advances in eye-tracking technology~\cite{Fuhl2016g}, three factors continue to adversely impact the performance of eye-tracking algorithms: 1) reflections from the surroundings and from intervening optics, 2) occlusions due to eyelashes, eyelid shape, or camera placement and 3) small shifts of the eye-tracker position caused due to slippage~\cite{Hansen2010b}. \rev{Pupil detection} algorithms such as ExCuSe~\cite{Fuhl2015a} and PuRe~\cite{Santini2018c} which rely on hand-crafted features are particularly susceptible to stray reflections (unanticipated patterns on eye imagery) and occlusion of descriptive gaze regions (such as eyelid covering the pupil or iris). Recent appearance-based methods based on Convolutional Neural Networks (CNNs) are better able to extract reasonably reliable gaze features despite the presence of reflections~\cite{chaudhary2019ritnet} or occlusions~\cite{Park2018e}. Additionally, for head-mounted eye-tracking systems, the degradation of gaze estimate accuracy over time due to slippage ~\cite{Kolakowski2005} can be minimized by estimating the 3D eyeball center of rotation~\cite{Santini2019a} (loosely referred at as an 'eyeball fit'). Estimating the precise physiology of the human eye is a complicated process and computationally intractable~\cite{Berard2016}. By making certain simplifying assumptions~\cite{Atchison2016b} about the human eye and its geometrical constraints, an estimate of a \textit{reduced} optical eyeball model can be obtained from 2D pupil~\cite{Swirski2012a,Swirski2013d,Kassner2014b,fuhl2020neural} or iris~\cite{Wood2014b,Wu2004,Plopski2015} elliptical fits. These elliptical fits are derived from identified pupil and iris segments or outline~\cite{fuhl2020tiny}. Efforts by Chaudhary \etal~\cite{chaudhary2019ritnet} and Wu \etal~\cite{Wu2019b} demonstrate that CNNs can precisely segment eye images into its constituent parts, \ie the pupil, iris, sclera and background skin regions. %(referred to as \textit{PartSeg} in this paper).

%However, ellipse fitting solutions produce degenerate or imprecise fits when pupil or iris regions are partially occluded.

%\sout{Park \etal observed that providing an auxiliary task of predicting occlusion-free ``gaze maps'' in remote eye imagery boosts the task of appearance-based gaze estimation~\cite{Park2018e}. It remains unexplored if a similar methodology can be applied to head-mounted eye-trackers in a computationally efficient manner while maintaining parity with higher precision requirements.}

%as it offers invariance to occlusion while enabling reliable elliptical fits in highly occluded eye images. 

%In this work, we show that ellipse fitting solutions produce degenerate or imprecise fits when pupil or iris regions are partially occluded. To mitigate this, we propose training CNNs using entire elliptical eye regions as opposed to visible eye-parts.
%Our proposed solution demonstrates robustness to occlusion while simultaneously producing accurate ellipse fits.

%Our proposed solution demonstrates robustness to occlusion while simultaneously producing accurate ellipse fits by modifying the standard eye-parts segmentation (PartSeg) into three categories, the full pupil, full iris, and background. 

%In this work, we show that ellipse fitting solutions produce degenerate or imprecise fits when pupil or iris regions are partially occluded.

\begin{figure}
    \centering
    \includegraphics[width=\linewidth]{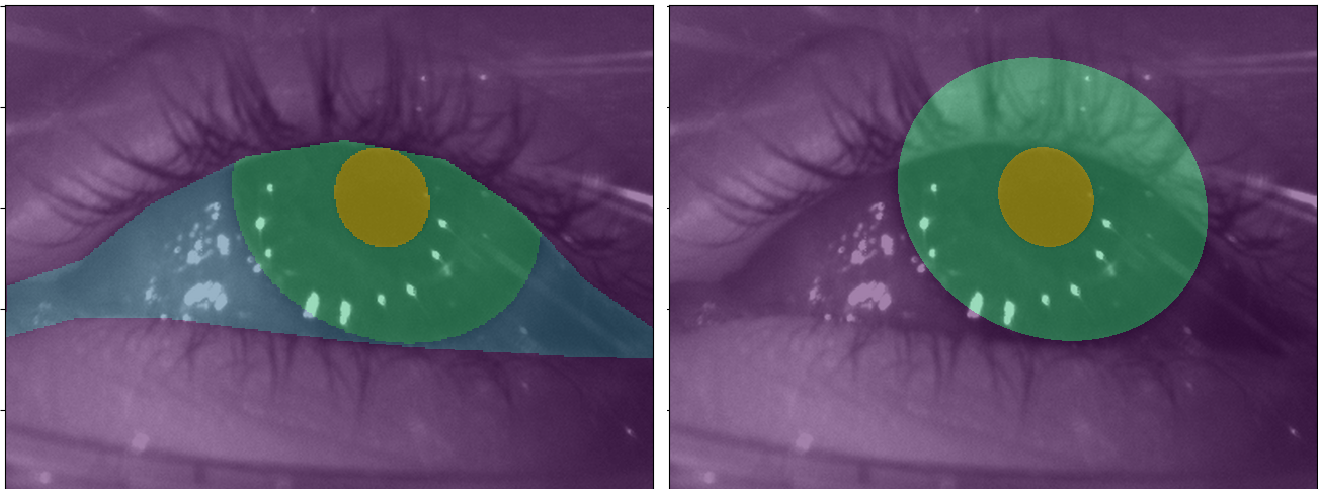}
    \caption{\textit{PartSeg vs EllSeg}. Left: A \textit{Four-class} eye part segmentation at the pixel-level (i.e. PartSeg) produces labelled pupil (yellow), iris (green), sclera (blue) and background (purple) classes. Right: The EllSeg (three-class) modification produces labelled pupil (yellow) and iris (green) elliptical regions and the rest is marked as background (purple).}

    % \caption{Comparison between PartSeg and EllSeg. Left: \textit{Four-class} eye parts segmentation (PartSeg). Right:  \textit{Three-class} ellipse segmentation (EllSeg). PartSeg produces labelled pupil (yellow), iris (green), sclera (blue) and background (purple) classes. EllSeg produces labelled pupil (yellow) and iris  (green) elliptical regions and the rest is marked as background (purple).}
    \label{fig:PartSeg_vs_EllSeg}
\end{figure}

In this work, we show that partially occluded pupil or iris regions can result in imprecise or degenerate elliptical fits. To mitigate this, we provide a solution, called \textit{EllSeg}, which is made robust to occlusion by training CNNs to predict entire elliptical eye regions (the full pupil and the full iris) along with the remaining background, as opposed to the standard visible eye-parts segmentation (PartSeg) (see Figure~\ref{fig:PartSeg_vs_EllSeg}). Additionally, we demonstrate that this approach enables us to train segmentation-based CNN architectures directly on datasets wherein only the pupil centers are available~\cite{Tonsen2016a,Fuhl2016e,Fuhl2017c}, allowing us to combine eye parts segmentation and pupil center estimation into a common framework.

%
%A drawback of relying solely on pupil ellipse stems from the fact that the pupil does not demonstrate elliptical behavior is large off axis conditions. 
%
%Recent preliminary work by Wang \etal leverage both iris and pupil ellipses to jointly estimate a reliable pupil center ($\sim$0.5 pixel reprojection error estimate)~\cite{Wang2019}. This approach shows promise and motivates the need for a joint limbus and pupil ellipse estimate.
The summary of our contributions are as follows:

\begin{enumerate}

    % \item We propose the EllSeg framework, a modification that can be utilized with any encoder-decoder segmentation framework for pupil and iris ellipse segmentation. EllSeg enables prediction of the pupil and iris as full elliptical structures despite the presence of occlusions.
    \item We propose EllSeg, a framework that can be utilized with any encoder-decoder architecture for pupil and iris ellipse segmentation. EllSeg enables prediction of the pupil and iris as full elliptical structures despite the presence of occlusions.

    % \item We propose the EllSeg framework, a full pupil and iris ellipse segmentation framework that allows a CNN to predict elliptical structures despite the presence of occlusions.
    
    %\item We show that the EllSeg framework enables estimating precise pupil and iris centers directly from predicted segmentation.
    
    % \sout{\item We produce improved ellipse fits as compared to the PartSeg framework, by jointly segmenting pupil and iris regions while estimating their centers in a single forward pass.}

    \item To establish the utility of our methodology, we rigorously test our proposed 3-class ellipse segmentation framework using three network architectures, a modified Dense Fully Connected Network~\cite{Jegou2017b} (referred as DenseElNet), RITnet~\cite{chaudhary2019ritnet} and DeepVOG~\cite{Yiu2019b}. Performance is benchmarked with well defined train and test splits on multiple datasets, including some which are limited to labelled pupil centers only.
    
    %The EllSeg formulation enables training segmentation networks on datasets with labelled pupil centers only.
\end{enumerate}

% EllSeg requires no additional computational overhead while providing resilience to occlusions and can be utilized with any encoder-decoder segmentation framework.

% \item Combine the task of predicting ellipse centers and EllSeg to unify the fields of pupil center-based tracking and eye image segmentation, enabling multi-set learning from a larger pool of datasets, which in turn provides access to a wide distribution of eye images. - \rk{This statement requires leave-one-out and full baseline table. If we don't have that then I'll remove.}

%  (\ac{@rakshit: i think we should mention parameters})
% \ck{reword this contribution, I was thrown off by the word compelling}
\section{Related work}
This work is primarily based on the observation that CNNs can identify which category a pixel belongs to despite conflicting appearance (\eg accurately predicting a pixel as belonging to the pupil despite being occluded by eyelids or glasses). Successful segmentation in the presence of ambiguous appearance indicates that a CNN can reason over a wide range of inter-pixel spatial relationships while precise segmentation boundaries indicate successful utilization of fine-grained, high-frequency content observed in local neighborhoods. This ability to capture local information with a global context is achieved by repeatedly pooling intermediate outputs of convolutional operations within a neural network~\cite{Dumoulin2016}. While numerous architectures can produce a ``one-to-one'' mapping between an image pixel and its segmentation output class, specific architectures rely on encoding an input image to low dimensional representation followed by decoding and up-sampling to a segmentation map - aptly named encoder-decoder architectures.

% One such network, introduced by Jegou \etal~\cite{Jegou2017b} proposed fully convolutional DenseNets (referred to as TiramisuNet in this paper) which demonstrated state of the art performance on multiple segmentation benchmark datasets. 
Researchers have demonstrated promising results using encoder-decoder architectures for image segmentation. For example, Chaudhary \etal~\cite{chaudhary2019ritnet} proposed RITnet, a lightweight architecture which leverages feature reuse and fixed channel size to maintain low model complexity while demonstrating state of the art performance on the OpenEDS dataset~\cite{Garbin2019d}. In this work, we designed our own encoder-decoder architecture called \textit{DenseElNet} which incorporates the dense block proposed by RITnet while leveraging residual connections across each block as proposed by Jegou \etal~\cite{Jegou2017b}. This ensures a healthy gradient flow and faster convergence while mitigating the vanishing gradient problem~\cite{He2015b,He2016a}. Similar to common encoder-decoder architectures, DenseElNet reduces the spatial extent of its input image but increases the channel size. Note that DenseElNet does not offer any particular novelty over existing encoder-decoder architectures. It is simply being used to facilitate testing of our EllSeg framework.

% \sout{Note that we do not claim any novelty on CNN architecture design and that the EllSeg framework can be applied to any encoder-decoder framework.} \ac{Note that our novelty is EllSeg framework instead of novel enocder-decoder architecture.}

The primary purpose of eye image segmentation, in the context of gaze estimation, is to produce reliable ellipse fits. The DeepVOG framework by Yiu \etal~\cite{Ronneberger2015b} utilizes the U-net architecture to segment the pupil followed by an out-of-network ellipse fitting procedure to generate a 3D model using the "two circles" approach~\cite{Swirski2013d,Safaee-Rad1992b}. A limitation of their approach is that they segment the pupil based solely on appearance which would likely suffer from occlusion as described previously. Fuhl \etal~\cite{Fuhl2019a} demonstrated that ellipse parameters can be regressed using the bottleneck representation of an input image. However they do not report any metrics for ellipse fit quality. Wu \etal~\cite{Wu2019b} leverage multiple decoders to segment an image and estimate 2D cornea and pupil center. Multiple decoders may increase computational requirements and introduce bottlenecks in the pipeline by operating on redundant information. In contrast, we show that the iris and pupil ellipse can be generated using a single encoder-decoder forward pass.

\section{Methodology}
\begin{figure}
    \centering
    \includegraphics[width=\linewidth]{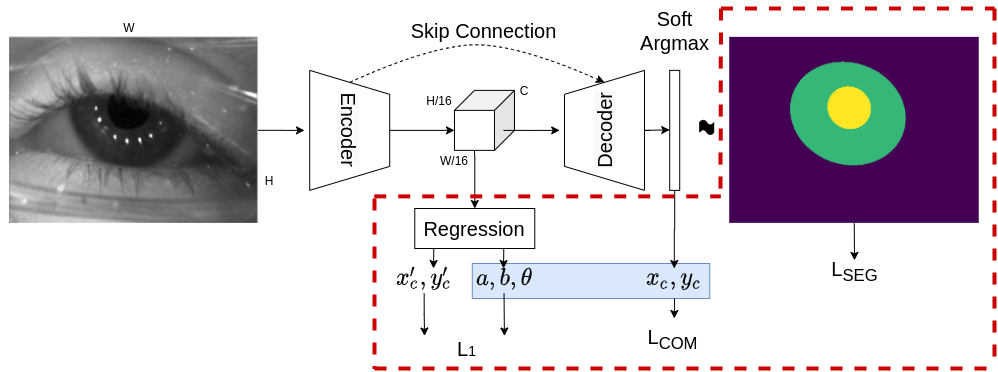}
    \caption{Proposed EllSeg framework (region enclosed by red dotted line) builds upon existing CNN-based approaches to facilitate the simultaneous segmentation and ellipse prediction for both iris and pupil regions. The resulting ellipse parameters are highlighted in the blue box.}
    \label{fig:EllSeg_framework}
\end{figure}

\begin{figure}
    \centering
    \includegraphics[width=\linewidth]{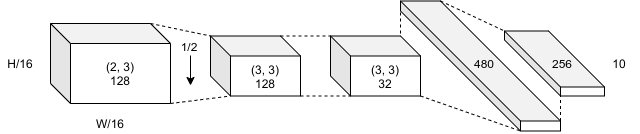}
    \caption{Regression module architecture. The $\downarrow$ signifies average pooling to $\nicefrac{1}{2}$ the resolution. Tensors are flattened after three convolutional layers and passed through two linear layers before regressing 10 values (5 ellipse parameters for pupil and iris each).}
    \label{fig:regress_mod}
\end{figure}

Figure~\ref{fig:EllSeg_framework} highlights the EllSeg framework on any generic encoder-decoder (E-D) architecture. First, an input image $I\subset \mathbb{R}$ is passed through an encoder to produce a bottleneck representation $Z$ such that $Z = E(I)$. In our implementation of DenseElNet, $I$ is down-sampled four times by a factor 2 at the bottleneck layer. Subsequently, the network segmentation output $O$ is given by $O = D(Z)$ and consists of three channels (background $O_{bg}$, iris $O_{ir}$ and pupil $O_{pl}$ output maps). Note that the segmentation outputs are also used to derive pupil and iris ellipse centers. The pupil and iris centers, along with the remaining ellipse parameters (axes and orientation), are also regressed from this bottleneck representation $Z$ using a series of convolutional layers followed by a flattening operation and mapped to a ten-dimensional output (5 parameters for both the iris and pupil ellipses). Please refer to Figure~\ref{fig:regress_mod} for the ellipse regression module architecture. We test the effectiveness of EllSeg framework on three architectures, DenseElNet (2.18M parameters), RITnet (0.25M parameters), and DeepVOG (3.71M parameters). Note that the regression module is trained alongside the entire network in an End-to-end fashion.

\subsection{Ellipse center}
The center of any convex shape can be described as a weighted summation of its spatial extent (see Equation~\ref{eq:COM}). In this context, \textit{spatial extent} refers to all possible pixel coordinates while \textit{weight} refers to the probability estimate of a pixel being within the convex structure.
\begin{equation}
    x^k_c, y^k_c = \sum_{i=1}^{W} \sum_{j=1}^{H} p^k_{<i,j>} x, \quad \sum_{i=1}^{W} \sum_{j=1}^{H} p^k_{<i,j>} y, \quad \quad p^k_{<i,j>} \subset \mathbb{R}
    \label{eq:COM}
\end{equation}
Here, $x^k_c$ and $y^k_c$ correspond to the center of a particular feature class $k$ (such as pupil). The iterators $i$ and $j$ span across the width $W$ and height $H$ of an image. The probability values $p^k$ for each pixel are derived after a scaled, spatial softmax operation~\cite{Nibali2018a}:

\begin{equation}
    p^k = \frac{\mathrm{exp}(\beta O_{<i,j>}^k)}{\sum_{i,j=1}^{W,H}\mathrm{exp}(\beta O_{<i,j>}^ k)}
    \label{eq:softargmax}
\end{equation}

Here, $\beta$ is a control parameter (also known as temperature~\cite{eriba2019kornia}), which scales network output around the largest value. We empirically set $\beta$ as 4. This formulation of ellipse center gives rise to several advantages offered by EllSeg over PartSeg discussed in Section~\ref{sec:Lcom} and Section~\ref{sec:improve_ellipse_estimate}.

While one may trivially estimate the pupil center in this manner, deriving the iris center is not straightforward due to its placement \textit{within} the pupil. One alternative is to sum the pupil and iris activation maps before spatial softmax. However, this incorrectly results in the predicted pupil and iris sharing the same 2D center which is physiologically improbable as the pupil is not usually perfectly centered within the iris~\cite{mosquera2015centration}. Instead, we propose leveraging the background class to predict the iris center in our 3 class segmentation framework. Encoder-decoder architectures have shown to perform exceedingly well at identifying "background" class pixels (see Supplementary Table 1 in Nair \etal~\cite{nair2020rit} and Table 2 in Wu \etal~\cite{Wu2019b}). To derive the iris center, we negate the background class output map in Equation~\ref{eq:softargmax}, a modification which subsequently leads to an inverted peak at the predicted iris center location. This inversion ensures the background probability scores do not affect segmentation based loss functions (see Section~\ref{sec:activation_name}).

%\ac{we should discuss}
%\gd{The bottleneck representation $Z$ can be thought of as a low-dimensional representation of the input image that contains the information necessary to perform the task for which the network has been trained . In the present context, this representation is expected to include a low-resolution segmentation of the eye into pupil, iris, and background pixels.}

% \gd{upon the pre-segmented pupil and iris}
\subsection{Ellipse axis and orientation}
 The bottleneck representation $Z$ is a low dimensional latent representation of the input image. This convenient representation enables us to regress parameters such as the ellipse axis and orientation (we use $L_1$ loss in our implementation). Experiments revealed that regressing the pupil and iris centers does not offer sub-pixel accuracy (see Section~\ref{sec:center_bottlevssoftargmax}) as opposed to deriving them from segmentation output as described in the previous section.

\subsection{Loss functions}
\subsubsection{Segmentation losses $\mathcal{L}_{SEG}$}
In the EllSeg framework, the network output $O$ is primarily used to segment an eye image into pupil and iris ellipses, and the background (which includes scleral regions). To train such an architecture, we use the combination of loss functions proposed in RITnet~\cite{Chaudhary2019c}. This strategy involves using a weighted combination of four loss functions; cross-entropy loss, $\mathcal{L}_{CEL}$, generalized dice loss~\cite{Sudre2017b} $\mathcal{L}_{GDL}$, boundary aware loss $\mathcal{L}_{BAL}$ and surface loss~\cite{Kervadec2018b} $\mathcal{L}_{SL}$.

The total loss $\mathcal{L}$ is given by a weighted combination of these losses as  $\mathcal{L}_{SEG} = \mathcal{L}_{CEL} (\lambda_1 + \lambda_2 \mathcal{L}_{BAL} ) + \lambda_3 \mathcal{L}_{GDL} + \lambda_4 \mathcal{L}_{SL}$. In our experiments, we used $\lambda_1 = 1$, $\lambda_2 = 20$, $\lambda_3 = (1 - \alpha)$ and $\lambda_4 = $ $\alpha$, where $\alpha = epoch/M$ and $M$ is the number of epochs.

\subsubsection{Center of Mass loss $\mathcal{L}_{COM}$}
\label{sec:Lcom}
The $\mathrm{L}_1$ loss function is used to formulate an error function between the center of mass, \ie the pupil and iris ellipse centers from the segmentation output maps, to their respective ground-truth centers. This enables us to leverage datasets such as \rev{ExCuSe~\cite{Fuhl2015a}}, ElSe~\cite{Fuhl2016e}, PupilNet~\cite{Fuhl2017c} and LPW~\cite{Tonsen2016a} in a segmentation framework where only the ground-truth pupil center is available. Note that COM $\mathrm{L}_1$ loss (henceforth referred to as $\mathcal{L}_{COM}$ loss) does not impede segmentation loss functions, but instead conditions the network output to jointly satisfy all loss functions. This results in the characteristic peaks observed in Section~\ref{sec:activation_name}. The inversion of the background class results in an inverted peak at the iris center location.

\section{Datasets}
\label{sec:dataset_prep}
Combining segmentation and $\mathcal{L}_{COM}$ losses allows the EllSeg framework to train CNNs on a large number of datasets (to the best of our knowledge, it enables the inclusion of all publicly available near-eye datasets). To demonstrate the utility of EllSeg, we choose the following datasets for our experiments: NVGaze~\cite{Kim2019b}, OpenEDS~\cite{Garbin2019d}, RITEyes, ElSe~\cite{Fuhl2016e}, ExCuSe~\cite{Fuhl2015a}, PupilNet~\cite{Fuhl2016f} and LPW~\cite{Tonsen2016a}. \rev{The ElSe and ExCuSe datasets are combined (also referred to as Fuhl) due to similar environment and eyetracker}. For more details about each dataset, available ground-truth modality, and train/test splits, please refer to Table~\ref{tbl:existing_datasets}. Note that we specifically leverage the S-General dataset from the RIT-Eyes framework ~\cite{nair2020rit} as it offers wide spatial distribution of \textit{eye} camera position.
% \begin{tabular}[c]{@{}c@{}}ElSe\\\rev{+ExCuSe}\end{tabular}
\begin{table*}
\centering
{%
\caption{Summary of datasets. $\uparrow$ and $\downarrow$ correspond to up and down sampling respectively. OpenEDS image crops are extracted around the scleral center followed by up-sampling. Note that images without valid pupil and iris fits are discarded (see Section~\ref{sec:dataset_prep}).}
\label{tbl:existing_datasets}
\begin{tabular}{|c|c|c|c|c|c|c|}
\hline
Dataset &
  Resolution &
  Train subset &
  Test subset &
  \begin{tabular} [c]{@{}c@{}} Groundtruth \\ included \end{tabular} &
  \begin{tabular}[c]{@{}c@{}}Image Count\\ (train, test)\end{tabular} &
  Preprocessing\\ \hline
NVGaze~\cite{Kim2019b} &
  1280$\times$960 &
  \begin{tabular}[c]{@{}c@{}}male 1-4\\ female 1-4\end{tabular} &
  \begin{tabular}[c]{@{}c@{}}male 5\\ female 5\end{tabular} &
  All &
  15623, 3895 &
  $\downarrow$4 \\
OpenEDS$^{19}$~\cite{Garbin2019d} &
  400$\times$640 &
  \begin{tabular}[c]{@{}c@{}}OpenEDS$^{19}$\\ train\end{tabular} &
  \begin{tabular}[c]{@{}c@{}}OpenEDS$^{19}$\\ valid\end{tabular} &
  PartSeg &
  8826, 2376 &
  \begin{tabular}[c]{@{}c@{}}Crop to\\ 400$\times$300\\ $\uparrow$1.6\end{tabular} \\
\begin{tabular}[c]{@{}c@{}}RITEyes\\ General~\cite{nair2020rit}\end{tabular} &
  640$\times$480 &
  Avatars 1-18 &
  Avatars 19-24 &
  All &
  33997, 11519 &
  $\downarrow$2 \\
LPW~\cite{Tonsen2016a}& 
  640$\times$480 &
  Subjects 1-16 &
  Subjects 17-22 &
  \begin{tabular}[c]{@{}c@{}}Pupil\\ center\end{tabular} &
  93127, 33388 &
  $\downarrow$2 \\
\rev{Fuhl~\cite{Fuhl2015a,Fuhl2016e}} &
  384$\times$288 &
  \begin{tabular}[c]{@{}c@{}}I, III, VI, VIII, IX,\\ XI, XIII, XV, XVII,\\ XIX, XX, XXII\end{tabular} &
  \begin{tabular}[c]{@{}c@{}}II, IV, V, VII,\\ X, XII, XIV, XVI\\ XVIII, XXI, XXIII\end{tabular} &
  \begin{tabular}[c]{@{}c@{}}Pupil\\ center\end{tabular} &
  60079, 33846 &
  $\uparrow$5/3 \\
PupilNet~\cite{Fuhl2016f} &
  384$\times$288 &
  I, III, V &
  II, IV &
  \begin{tabular}[c]{@{}c@{}}Pupil\\ center\end{tabular} &
  25471, 15707 &
  $\uparrow$5/3 \\ \hline
\end{tabular}%
}
\end{table*}

\subsection{Groundtruth ellipse fits}
\label{sec:groundtruthEllipseFits}
\begin{figure}
    \centering
    \includegraphics[width=\linewidth]{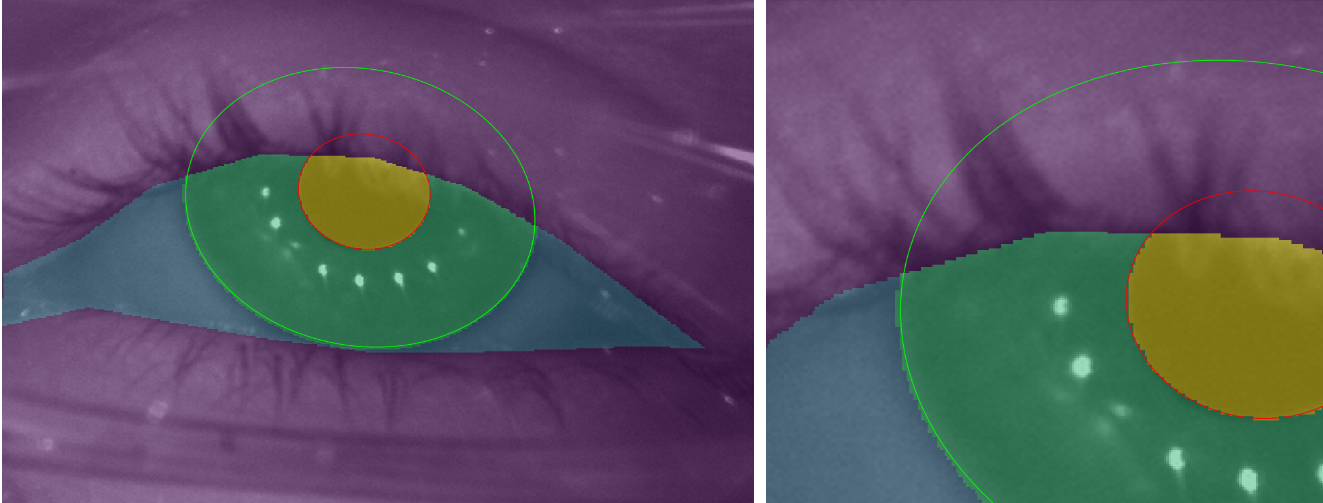}
    \caption{Ellipse fitting quality on ground truth PartSeg masks. These fits are further used to generate EllSeg masks for the OpenEDS dataset.}
    \label{fig:ellipse_fit}
\end{figure}

To obtain groundtruth pupil and iris ellipse fits from the selected datasets, pupil and limbus edges are extracted from groundtruth segmentation masks using a canny edge detector. To ensure subpixel accuracy, we consider edge pixels in the inverted mask as well. Edge pixels which satisfy pupil-iris (\ie no neighboring sclera or background pixel) or limbus (\ie no neighboring pupil or background pixel) conditions are used to determine ellipse parameters using the ElliFit algorithm~\cite{Prasad2013a} (see Figure~\ref{fig:ellipse_fit}). Random Sample Consensus (RANSAC)~\cite{fischler1981random} is employed to remove outliers \rev{with residuals higher than $5\times10^{-3}$, an empirically derived threshold}. While datasets such as RITEyes and NVGaze directly offer EllSeg compatible groundtruth semantic masks, synthetic masks for OpenEDS were generated based on elliptical fits. Images without valid pupil or iris fits (117 out of 11319) were discarded from all subsequent analysis.

%\gd{Within the context of ellipse-fitting, images are herafter referred to as \textbf{valid} or \textbf{invalid} on the basis of whether they contain the minimum of five non co-linear points necessary to fit an ellipse.  Of the X images in the original OpenEDS dataset, Y were valid, and Z invalid. }

\section{Experiments and Hypothesis}
\begin{figure}
\begin{center}
\includegraphics[width=\linewidth]{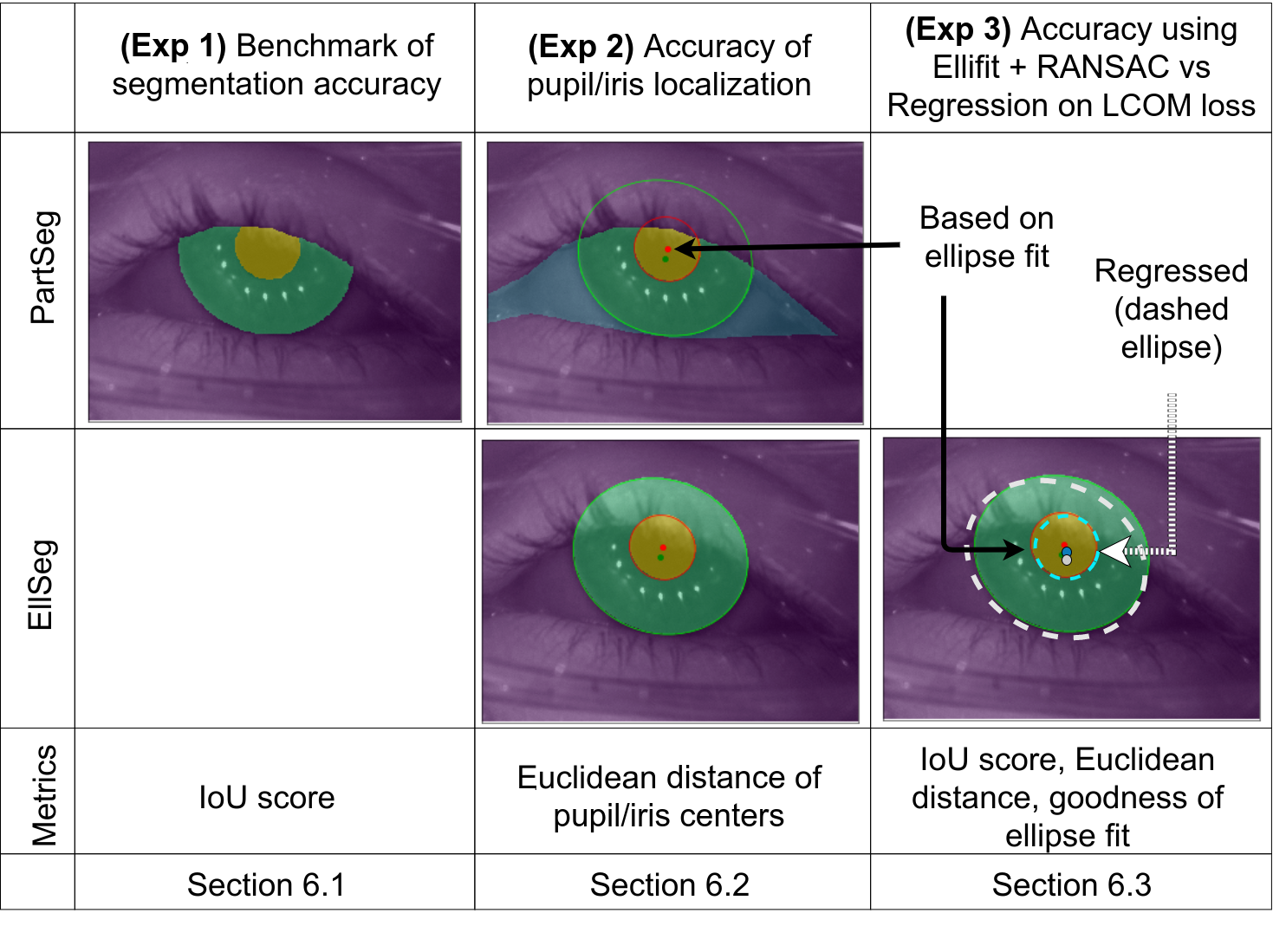}
\end{center}
\caption{Summary of all experiments described in following sections (Center estimates are best viewed on screen).}
\label{fig:hypothesis}
\end{figure}

We rigorously test various hypotheses to validate the efficacy of our proposed methodology in the field of eye-tracking. In the first experiment (Section~\ref{sec:state_of_art_models}), we benchmark the segmentation performance of our network, DenseElNet, on the standard PartSeg framework. Comparable or superior performance on the PartSeg task will validate DenseElNet. In the second experiment (Section~\ref{sec:center_estimate}), we test whether the EllSeg framework improves the detection of both pupil and iris estimates over its PartSeg counterpart. Finally, in the third experiment (Section~\ref{sec:improve_ellipse_estimate}), we compare the results of regressing elliptical parameters in the EllSeg framework to those found when estimating the ellipse parameters using RANSAC. This experiment will test whether reliable and differentiable ellipses can be directly estimated in an encoder-decoder architecture. Summary of all the experiments can be found in Figure~\ref{fig:hypothesis}.

\subsection{Training}
To ensure fair comparison, all CNN architectures are trained and evaluated with identical train/validation/test splits. The \rev{development} set is divided into a 80/20$\%$ train/validation split. Sample selection is stratified based on binned 2D pupil center position and subsets present within each dataset (see Table~\ref{tbl:existing_datasets}). This approach ensures that biases introduced due to sampling are minimized while maintaining similar statistical distributions across training and validation sets. Bins with fewer than five images are automatically discarded. All architectures are trained using ADAM optimization~\cite{Kingma2015} on a batch of 48 images at 320x240 resolution with a learning rate of $5\times10^{-4}$ on an NVIDIA V100 GPU.

During training, all models were evaluated with the metric: $[4 + \mathrm{mIoU} - 0.0025(d_p + d_i) -  \nicefrac{(\theta_p + \theta_i)}{90^\circ}]$, where mIoU corresponds to the mean intersection over union (IoU)~\cite{everingham20052005} score which quantifies segmentation performance, $d_p$ \& $d_i$ are the distances between pupil and iris centers from their groundtruth values in pixels, and $\theta_p$ \& $\theta_i$ are the angular error between the predicted and groundtruth ellipse orientations in degrees. If no improvement above $10^{-3}$ was observed on this metric for ten consecutive epochs, then a network's parameters were deemed converged. The learning rate was reduced by a factor of ten if no improvements were identified over five epochs. To reduce training time and ensure stable training on pupil-center-only datasets, all models were pretrained on NVGaze, OpenEDS and RIT-Eyes training sets for two epochs.

\subsection{Data augmentation}
To increase the robustness of models and avoid overfitting, training images were randomly augmented with the following procedures with equal probability (12.5\%) of occurrence:
\begin{itemize}
    \item Horizontal flips
    \item Image rotation up to $\pm$30$^\circ$
    \item Addition of Gaussian blur with 2$\leq\sigma\leq$7
    \item Random Gamma correction for $\gamma=$[0.6, 0.8, 1.2, 1.4]
    \item Exposure offset up to $\pm$25 levels
    \item Gaussian noise with 2$\leq\sigma\leq$16
    \item Image corruption by masking out pixels along a four-pixel thick line
    \item No augmentation
\end{itemize}

\subsection{Evaluation Metrics}
\label{sec:evalMetrics}
All segmentation performance is evaluated by IoU scores. Ellipse center accuracy is reported as the Euclidean distance in pixel error from their respective groundtruth annotations. Additionally, pupil and iris detection rate~\cite{Swirski2012a}, \ie the percentage of ellipse centers accurately identified within a range of pixels of the groundtruth center point is also reported.
%Note that this formulation of detection rate is heavily dependant on the resolution of eye imagery, eyetracker placement and system design.

\begin{figure}[h]
\begin{center}
\includegraphics[width=\linewidth]{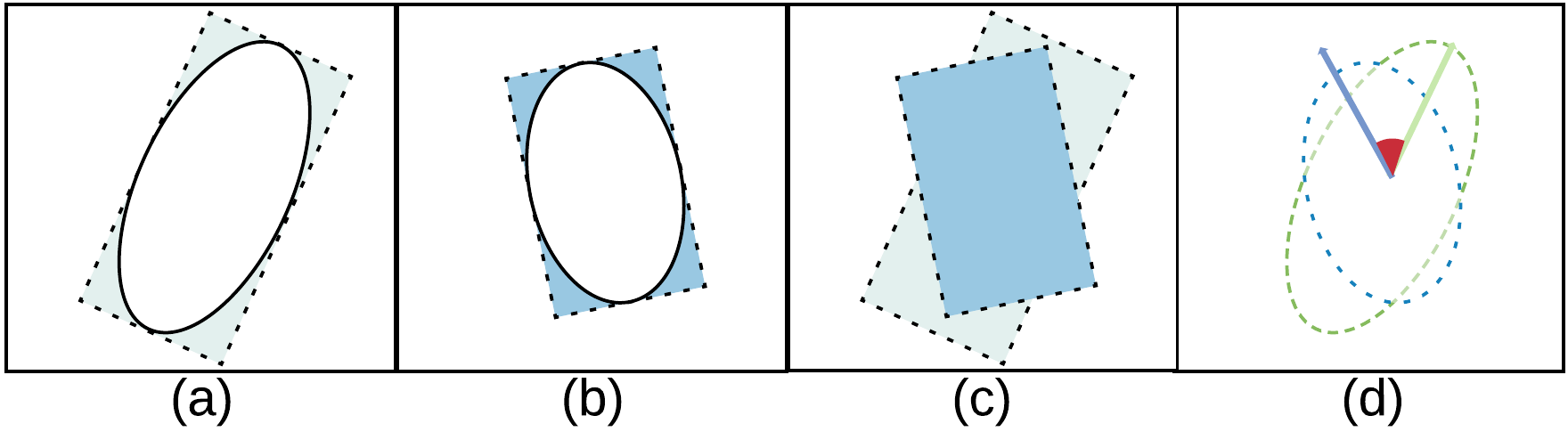}
\end{center}
\caption{Visualization of goodness of fit metrics used in the paper. (\textbf{a}) Groudtruth ellipse (pupil or iris). (\textbf{b}) Corresponding predicted ellipse. The rectangular boxes denote ellipse-axis-aligned bounding boxes for the respective ellipses. (\textbf{c}) denotes the bounding box overlap region and (\textbf{d}) illustrates the angular difference between the two ellipses.}
\label{fig:ellipse_explanation}
\end{figure}

As most gaze estimation algorithms rely on ellipse fitting on the segmented pupil and/or iris, we quantify elliptical goodness of fit with metrics that effectively capture ellipse offset, orientation errors and scaling errors. In this work, we utilize a bounding box overlap IoU metric that accounts for all ellipse parameters: center, axes, and orientation. For each defined elliptical structure, a enclosing bounding box is generated. IoU scores are obtained from a comparison between groundtruth and predicted bounding boxes (Figure~\ref{fig:ellipse_explanation}). Note that the orientation error (difference in ellipse orientation) of the fits is calculated for images in which the ratio of major to minor axis length exceeded 1.1 - this avoids large artifacts when elliptical fits are nearly circular.

% The measure of boundary IOU relies upon a subsequent fit of a two bounding boxes - one fit to the fit ellipse, and one fit to the ground truth. These bounding boxes were then used to calculate an IoU score that reflects the amount of overlap, where an IoU of 1 would reflect a perfect overlap, and thus a perfect fit. The orientation error of the fits was also calculated for images in which the ratio of major to minor axis length exceeded 1.1 - a step intended to avoid large artifacts in nearly circular cases. 

\section{Results and Discussion}
\subsection{Comparison with state-of-the-art models}
\label{sec:state_of_art_models}

The DenseElNet architecture is a hybrid of RITnet and TiramisuNet, and has 2.18M parameters. We also explore the alternative possibility of utilizing other state-of-the-art encoder-decoder architectures like DeepVOG and RITNet. DeepVOG, with 3.71M parameters, segments images into two classes; \textit{pupil} and \textit{background}, \ie (non-pupil). RITnet, with 0.25M parameters, defines four classes; \textit{pupil}, \textit{iris}, \textit{sclera}, and \textit{background}  (other). Table~\ref{tab:compare_network_archs} highlights that both RITnet and DenseElNet models outperform DeepVOG on every dataset. Table~\ref{tab:compare_network_archs} also demonstrates that the performance of DenseElNet and RITnet are comparable ($<2\%$ difference) on all datasets despite varying model complexity.

\begin{table}[h]
% table caption is above the table
\centering
\renewcommand{\arraystretch}{1.25}
\caption{Eye Parts Segmentation: Comparison of \textit{pupil} (and \textit{iris}, inside parenthesis) \textit{class} IoU scores for RITnet, DeepVOG and DenseElNet model architectures (along rows) in OpenEDS, NVGaze and RIT-Eyes dataset (along columns). Bold values indicate the best performance within each dataset. Because DeepVOG was not trained to segment the iris, we are unable to provide iris IOU scores.}
\label{tab:compare_network_archs}       % Give a unique label
\begin{tabular}{|c|c|c|c|}
\hline
 Model & OpenEDS & NVGaze & RIT-Eyes  \\
  %& \textbf{Previous work} & \textbf{Current work}\\
\hline
RITnet & 95.0 (91.4) & \textbf{93.2 (91.7)} & 89.5/94.4\\
\hline
DeepVOG & 89.1 (NA)  & 90.9 (NA)  & 83.5 (NA) \\
\hline
    DenseElNet & \textbf{95.4 (92.1)} & 93.1 (91.4) &\textbf{91.5 (95.4)}\\
\hline
\end{tabular}
\end{table}

\subsection{Ellipse center estimation}
\label{sec:center_estimate}
In this section, we explore the usefulness of the full ellipse segmentation (EllSeg) over the traditional eye parts segmentation (PartSeg) by comparing the pupil/iris center detection rates. We train three network architectures; RITnet, DeepVOG, and DenseElNet both with $\mathcal{L}_{SEG}$ loss functions using the following training scenarios: 

\begin{itemize}
\item Traditional, four class PartSeg (referred as \textit{RITnet-PartSeg},
\textit{DeepVOG-PartSeg}, and \textit{DenseElNet-PartSeg})
\item 3-class EllSeg (referred as \textit{RITnet-EllSeg}, \textit{DeepVOG-EllSeg}, and \textit{DenseElNet-EllSeg})
\end{itemize}

%^Both scenarios utilize the $\mathcal{L}_{SEG}$ loss functions. 
Note that, in this section, all ellipse centers are derived by utilizing ElliFit~\cite{Prasad2013a} along with RANSAC outlier removal on output segmentation maps.

\begin{figure*}
\begin{center}
\includegraphics[width=\linewidth]{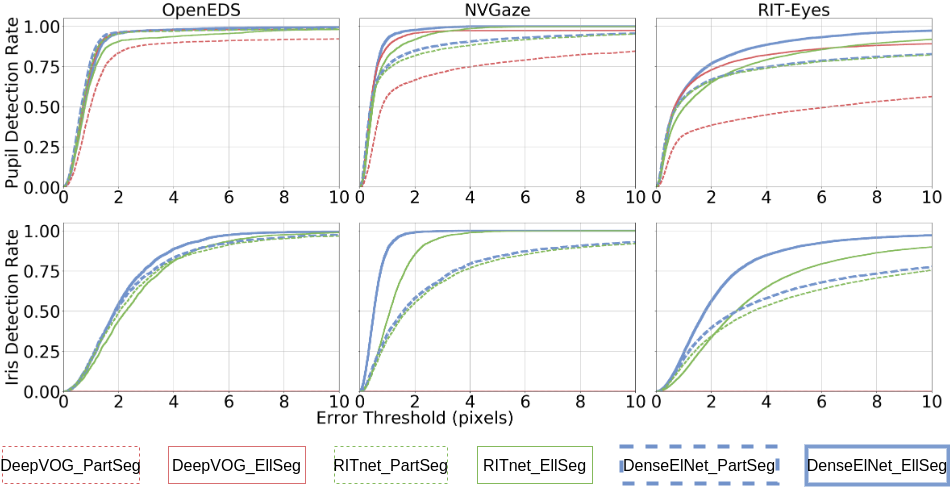}
\end{center}
\caption{\textit{PartSeg vs EllSeg}: The pupil detection rate (top row) and iris detection rate (bottom row) as a function of the threshold for tolerated pixel error for center approximation for OpenEDS (left column), NVGaze (middle column) and RIT-Eyes (right column). Results for three architectures RITnet, DeepVOG and DenseElNet are present for both cases PartSeg (dashed lines) and EllSeg (solid lines). Note that only the pupil detection rate is shown for the DeepVOG architecture. All detection rates presented here are derived using ellipse fits on segmentation outputs on images sized at 320 $\times$ 240. Here, one pixel error corresponds to 0.25\% of the image diagonal length.}
\label{fig:pupil_c_PartSeg-EllSeg}
\end{figure*}

Figure~\ref{fig:pupil_c_PartSeg-EllSeg} presents the pupil/iris detection rate as a function of the error threshold (in pixels) for DeepVOG, RITnet, and DenseElNet, using both PartSeg and EllSeg frameworks. Although all models demonstrate similar performance when tested upon the OpenEDS dataset, models trained using the EllSeg framework demonstrate superior pupil and iris detection on the NVGaze and RIT-Eyes datasets. 

Analysis of the ground truth imagery suggests that this difference may be attributed to the varying amounts of pupil/iris occlusion within each dataset. In order to verify this, we compute \textit{occlusion magnitude}, $O_m$, which is defined as one minus the IoU of PartSeg and EllSeg ground truth maps. Based on this magnitude, each image is classified into 3 categories of occlusion (shown in Table~\ref{tab:compare_occlusion})  based on empirical thresholds, a) fully occluded ($O_m \geq 0.7$) b) partially occluded ($0.3 \leq O_m < 0.7$) and c) fully visible ($O_m < 0.3$). 

\begin{table}
\caption{The percentage of images classified as three categories of occlusion (see Section~\ref{sec:center_estimate}) for each dataset. Values are presented as pupil (iris).}
\label{tab:compare_occlusion}
\centering
\renewcommand{\arraystretch}{1.25}
\begin{tabular}{|c|c|c|c|}
\hline
        & Occluded   & Partial     & Visible     \\ \hline
OpenEDS & 0.0 (0.0)  & 1.5 (17.2)  & 98.5 (82.7) \\ \hline
NVGaze  & 2.3 (0.0)  & 14.8 (75.6) & 82.9 (24.4) \\ \hline
RITEyes & 9.5 (11.1) & 70.7 (22.3) & 19.8 (66.7) \\ \hline
\end{tabular}
\end{table}

%This analysis reveals that the pupil is partially occluded on 1.5\% of images in the OpenEDS dataset, while the iris is partially occluded on 17.2\% of imagery.

Dramatic improvements can be observed for the NVGaze and RITEyes datasets wherein a large percent of images demonstrate partially occluded iris or pupil. Since a smaller percent of images are occluded in the OpenEDS dataset, we observe a small but consistent improvement in the iris detection rate between 3-6 pixel error threshold (see Figure~\ref{fig:pupil_c_PartSeg-EllSeg}, second row-first column). These results and subsequent analysis clearly demonstrate that EllSeg is robust to occlusions.

In addition to improving ellipse center estimates, Table~\ref{tab:number_count} demonstrates that the EllSeg protocol reduces the number of images with invalid ellipse fits on the predicted segmentation output.

%Although there is a drastic improvement with the EllSeg framework for all model architectures, the largest improvement is seen in RIT-Eyes, which has a large number of images with varying degrees of eye-closure causing occlusion.

% , consistent with the valid/invalid distinction presented in section \ref{sec:groundtruthEllipseFits}

% Table~\ref{tab:number_count} presents the number of images that failed pupil and/or iris center detection, consistent with the valid/invalid distinction presented in section \ref{sec:groundtruthEllipseFits}. There is a drastic improvement with the EllSeg framework for all model architectures. The largest improvement is seen in RIT-Eyes, which has a large number of images with varying degrees of eye-closure causing occlusion.

% Further, Table~\ref{tab:number_count} highlights the number of images that failed pupil and/or iris center detection because of insufficient number of points on the border to fit an ellipse. We see a drastic improvement with the EllSeg framework for all model architectures. The largest improvement is seen in RIT-Eyes, which has a large number of images with varying degrees of eye-closure causing occlusion.

%Note that for cases of partial occlusion, ellipse fits are sometimes incorrect resulting in high pixel error. Some recent approach such as 

\begin{table}[h]
\centering
\renewcommand{\arraystretch}{1.25}
\caption{The number of images without valid PartSeg or EllSeg ellipse fits for pupil (and iris, inside parenthesis) for DeepVOG, RITnet, and DenseElNet. The total column represents the number of valid images used for testing (as in section \ref{sec:groundtruthEllipseFits}). Bold text (lower number) shows superior performance and illustrates the effectiveness of the EllSeg framework.}

% \caption{Number of images without valid PartSeg or EllSeg ellipse fits for pupil (and iris, inside parenthesis) for DeepVOG, RITnet, and DenseElNet. Only labels that passed ellipse fitting test on ground truth were used for testing (shown in the total column). Bold text (lower number) shows superior performance and illustrates the effectiveness of the EllSeg framework.}
\label{tab:number_count}       % Give a unique label
\begin{tabular}{|c|c|c|c|c|c|}
\hline
                     & Dataset    & Total & DeepVOG & RITnet   & DenseElNet \\ \hline
\multirow{3}{*}{\spheading{PartSeg}} & OpenEDS  & 2376  & 17 (NA)      & 1 (0)    & 2 (0)     \\ \cline{2-6} 
                         & NVGaze   & 3895  & 10 (NA)      & 0 (0)    & 0 (0)     \\ \cline{2-6} 
                         & RIT-Eyes & 11519 & 1072  (NA)   & 287 (69) & 353 (62)  \\ \hline
\multirow{3}{*}{\spheading{EllSeg}}  & OpenEDS  & 2376  & \textbf{6}  (NA)     & 1 (0)    & \textbf{0} (0)     \\ \cline{2-6} 
                         & NVGaze   & 3895  & \textbf{0} (NA)        & 0 (0)     & 0 (0)     \\ \cline{2-6} 
                         & RIT-Eyes & 11519 & \textbf{215} (NA)     & \textbf{60} (\textbf{18})  & \textbf{1} (\textbf{0})     \\ \hline
\end{tabular}
\end{table}

\subsection{Improving the ellipse estimates}
\label{sec:improve_ellipse_estimate}
In this section, we analyze the impact of $L_{COM}$ on segmentation output maps, ellipse shape parameters and ellipse center estimates. 

\begin{figure*}
\begin{center}
\includegraphics[width=\linewidth]{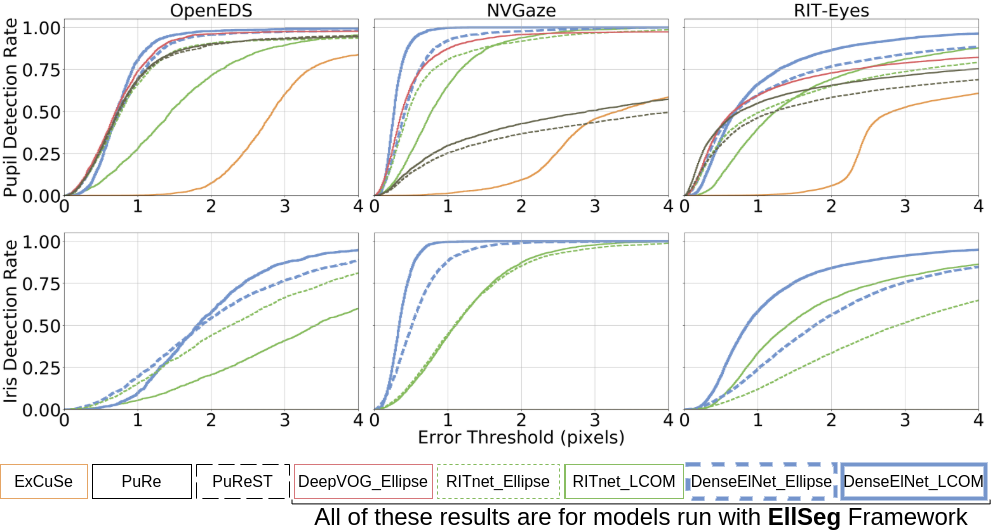}
\end{center}

\caption{\textit{EllSeg with and without $\mathcal{L}_{COM}$ loss}: The pupil detection rate (top row) and iris detection rate (bottom row) for various pixel error thresholds of center approximation for three datasets. Models (RITnet, DenseElNet and DeepVOG) are trained with the EllSeg framework before the pupil center is estimated using either the ElliFit segmentation output map, or with $\mathcal{L}_{COM}$ loss. The result for non-CNN based model ExCuSe, PuRe and PuReST are also shown. One pixel error corresponds to 0.25\% of the image diagonal length.}

% \caption{Similar to Figure~\ref{fig:pupil_c_PartSeg-EllSeg}, this figure shows the pupil detection rate (top row) and iris detection rate (bottom row) vs the pixel error in center approximation for three datasets. Here, models (RITnet, DenseElNet and DeepVOG) are trained \gd{with the EllSeg framework} \sout{for full-ellipse condition for two cases} and either a) with ellipse fits to predict the pupil/iris center and b) with $\mathcal{L}_{COM}$ loss. The result for non-CNN based model ExCuSe is also shown. Here, one pixel error corresponds to 0.25\% of the image diagonal length.}
\label{fig:Lcom_pupil}
\end{figure*}

Ellipse center estimates results are shown in Figure~\ref{fig:Lcom_pupil}. All models (RITnet, DeepVOG and DenseElNet) are trained with the EllSeg framework \textit{with} and \textit{without} $\mathcal{L}_{COM}$. Ellipse centers \textit{without} $\mathcal{L}_{COM}$ loss are estimated using ElliFit on segmentation output maps. Models trained \textit{with} $\mathcal{L}_{COM}$ loss estimate their centers ($x_{c}$ and $y_{c}$) as shown in Figure~\ref{fig:EllSeg_framework}.

%Note that these generated by the network directly and do not require an added ellipse fitting operation during testing.
 
%Like Figure~\ref{fig:pupil_c_PartSeg-EllSeg}, Figure~\ref{fig:Lcom_pupil} also compares pupil and iris detection rate with the corresponding pixel error.

Figure~\ref{fig:Lcom_pupil} also includes the results of non-CNN based algorithms ExCuSe~\cite{Fuhl2015a}, PuRe ~\cite{Santini2018c}, and PuReST ~\cite{Santini2018b} which rely on filtered edges, morphological operations and handcrafted features using computer-vision based methods. Note that none of these methods were designed for OpenEDS, NVGaze, or RITeyes datasets. To facilitate application, pixels with a ground truth label identifying them as a member of the "background" class are converted to a uniform grey (digital count$=$127). This step minimizes the chance of false detection of the pupil within the background, which is a common issue for images within the OpenEDS and NVGaze datasets, which have black regions in the periphery. Note that for ExCuSe, images are resized to the author-recommended size (384x288). The predicted center is then remapped to (320x240) to facilitate comparison. For PuRe and PuReST, the EyeRecTool~\cite{santini2017eyerectoo} is used to compute pupil center using the original image size (320x240).

% Figure~\ref{fig:Lcom_pupil} presents the pupil and iris detection rates as a function of the error threshold (pixels) for DeepVOG, RITnet, and DenseElNet with or without $\mathcal{L}_{COM}$ loss. 

Figure~\ref{fig:Lcom_pupil} reveals that, although introduction of $\mathcal{L}_{COM}$ often degraded the performance of RITnet, it improved performance for our model, (DenseElNet). Further, for pupil detection, the models trained using CNN outperforms all the non-CNNs based models ExCuSe, PuRe and PuReST.
 
% \sout{This performance improvement is most dramatic at pixel error threshold values greater than 0.5 pixels for the pupil, and 1 pixels for the iris center.}

Table~\ref{tab:compare_withLcom} shows the comparison of median values of pupil center estimates \textit{with} and \textit{without} $\mathcal{L}_{COM}$ loss in regards to both models RITnet and DenseElNet. There is a slight improvement in the median values in the DenseElNet model with the introduction of this loss function. However, for the RITnet model, the inclusion of $\mathcal{L}_{COM}$ deteriorated the performance by 57\%, 19\%, and 19\% for OpenEDS, NVGaze, and RIT-Eyes datasets respectively (within one-pixel error range for Pupil center). We suspect this behavior is due to the relatively limited channel size and low parameter count of RITnet when compared to DenseElNet.

\begin{table}
\centering
\renewcommand{\arraystretch}{1.25}
\caption{Comparison of Pupil center estimate errors (in pixels) on various datasets in terms of median scores. Note all the CNN models are trained with EllSeg framework. Image size is 320 $\times$ 240.}
\label{tab:compare_withLcom}       % Give a unique label
\begin{tabular}{|c|c|c|c|c|}
 \hline
 Model & \multicolumn{2}{c|}{RITnet} &   \multicolumn{2}{c|}{DenseElNet}  \\
 \hline
 Method & Ellipse fit &$\mathcal{L}_{COM}$ & Ellipse fit &$\mathcal{L}_{COM}$\\
\hline
OpenEDS & 0.8 &1.5& 0.8 &0.7\\
NVGaze & 0.5 &0.8& 0.4 &0.3\\
RIT-Eyes & 1.0 &1.2& 0.7 &0.7\\
Fuhl & - &73.4& - &1.7\\
LPW &  - &4.7& - &0.8\\
PupilNet  & - &77.6& - &1.6\\
\hline
\end{tabular}
\end{table}

\begin{figure}
\begin{center}
\includegraphics[width=\linewidth]{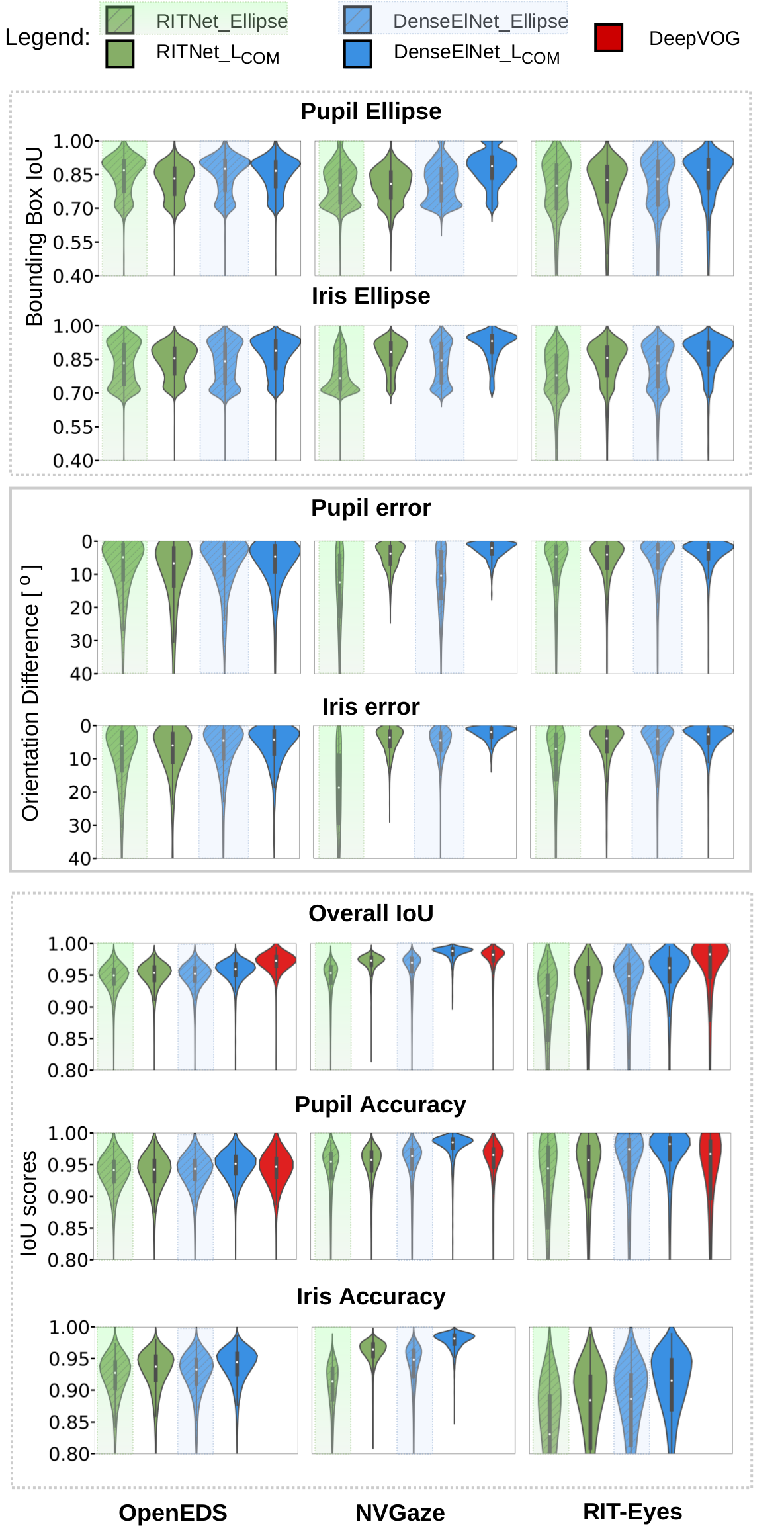}
\end{center}
\caption{ Violin plots of boundary overlap IoU (1st and 2nd row: top dashed box), orientation error (3rd and 4th row: middle solid box), and segmentation IoU score (last three rows: bottom dashed box) following EllSeg framework by  RITnet and DenseElNet, with or without $\mathcal{L}_{COM}$ loss ($\mathcal{L}_{COM}$ vs Ellipse), following application to the OpenEDS, NVGaze, and RIT-Eyes datasets (columns)} \textit{(Best viewed on screen)}.
\label{fig:ellipse_compare}
\end{figure}

\iffalse
\begin{figure}
\begin{center}
\includegraphics[width=\linewidth]{results/part_3_3.png}
\end{center}

\caption{The overall IoU, and IoU for pupil and iris fits following EllSeg framework by  RITnet and DenseElNet, with or without $\mathcal{L}_{COM}$ loss ($\mathcal{L}_{COM}$ vs Ellipse), following application to the OpenEDS, NVGaze, and RIT-Eyes datasets (rows) \textit{(Best viewed on screen)}.}
\label{fig:seg_score}
\end{figure}
\fi

The analyses presented up to this point focus on the accuracy of pupil/iris center estimates. However, many algorithms for gaze estimation rely on accurate estimation of pupil and iris ellipses for the construction of 3D geometric models of the oriented eye ~\cite{Yiu2019b,Kassner2014b,Swirski2013d,Wood2014b}. This necessitates a quantitative measure for the goodness of an ellipse fit. The methodology presented in Section~\ref{sec:evalMetrics} and represented in Figure~\ref{fig:ellipse_explanation} is used to calculate the \textit{boundary IOU} - a measure used to estimate the quality of boundary estimation. Boundary IoU was calculated for both the pupil and the iris after application of RITnet and Densenet to several datasets, either with or without $\mathcal{L}_{COM}$. When $\mathcal{L}_{COM}$ is used, ellipse orientation and axis parameters are regressed via the bottleneck layer, and when it is not, the ellipse is fit to the segmented mask. 

%When $\mathcal{L}_{COM}$ is used, ellipses are fit through a process of regression, and when it is not, the ellipse is fit to the segmented mask. 

The result of this analysis are presented in Figure~\ref{fig:ellipse_compare}, and reveal that that DenseElNet \textit{with} $\mathcal{L}_{COM}$ outperforms \textit{without} $\mathcal{L}_{COM}$ in terms of boundary IOU and orientation error for both, the pupil and iris, on almost all datasets.

% Whereas the analyses presented up to this point focus on the accuracy of estimations of the pupil/iris center, many algorithms for gaze estimation rely upon accurate segmentation of pupil and iris boundaries, for example, for the construction of 3D geometric models of the oriented eye ~\cite{Yiu2019b,Kassner2014b,Swirski2013d}. Figure~\ref{fig:ellipse_compare} presents the \textit{boundary IOU}, which is a measure of the quality of boundary estimation of pupil and iris fits to several datasets using RITnet and Densenet, and either with or without $\mathcal{L}_{COM}$. When $\mathcal{L}_{COM}$ was used, Ellipses were fit through a process of regression, and when it was not, the ellipse was fit to the segmented mask. The measure of boundary boundary IOU relies upon a subsequent fit of a two bounding boxes - one fit to the fit ellipse, and one fit to the ground truth. These bounding boxes were then used to calculate an IoU score that reflects the amount of overlap, where an IoU of 1 would reflect a perfect overlap, and thus a perfect fit. The orientation error of the fits was also calculated for images in which the ratio of major to minor axis length exceeded 1.1 - a step intended to avoid large artifacts in nearly circular cases. As presented in Figure~\ref{fig:ellipse_compare}, this analysis reveals that that DenseElNet \textit{with} $\mathcal{L}_{COM}$ outperforms in terms of boundary IOU and orientation error for both the pupil and iris cases on almost all datasets.

\begin{figure*}
\begin{center}
\includegraphics[width=\linewidth]{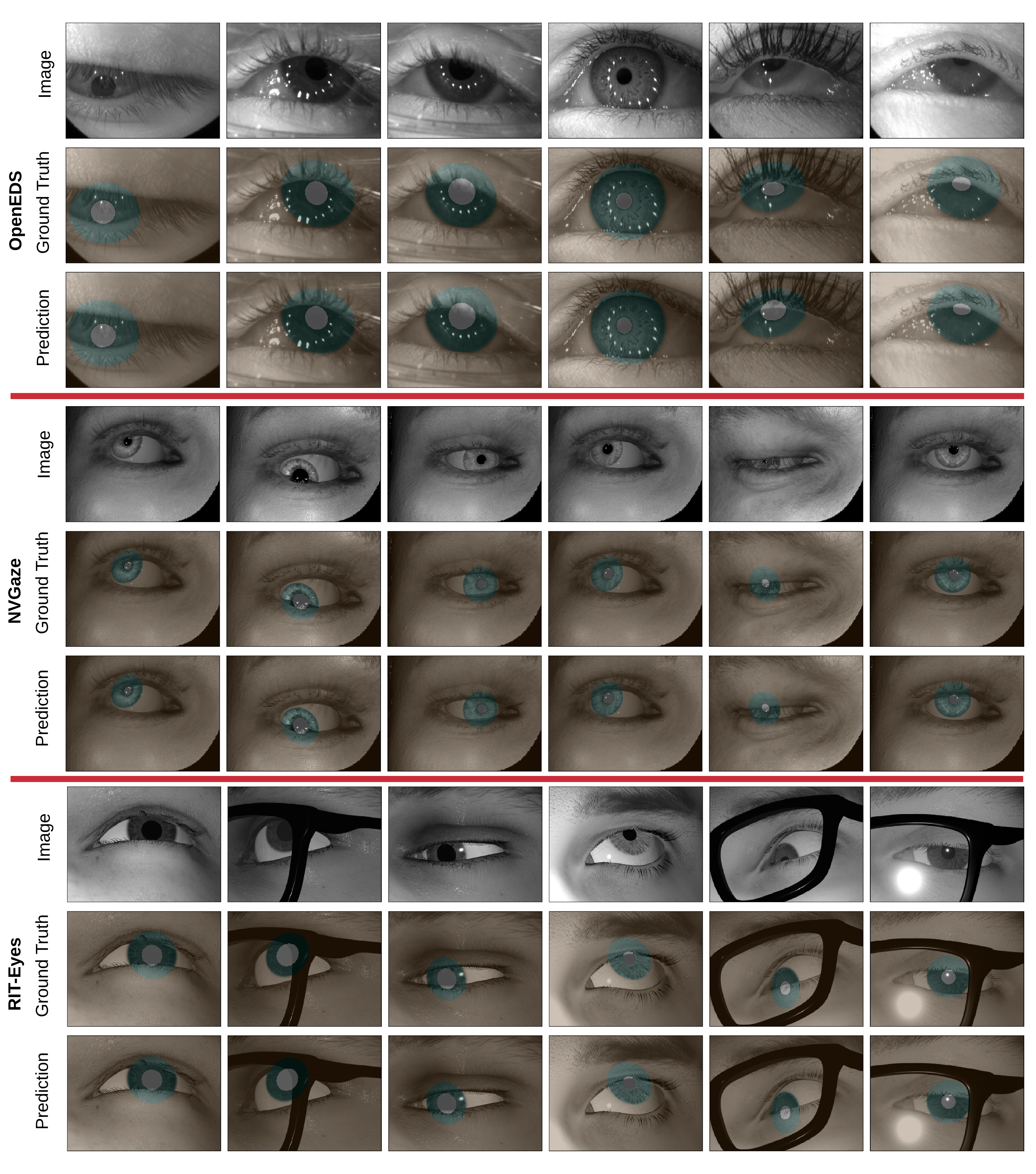}
\end{center}
\caption{DenseElNet model prediction and its respective ground truth for OpenEDS, NVGaze and RIT-Eyes dataset.}
\label{fig:ground_truth_prection}
\end{figure*}

The pixel-wise IOU score of iris and pupil segmentation is presented in Figure~\ref{fig:ellipse_compare} (last three rows). This analysis reveals that DenseElNet also outperforms other models in the segmentation of the pupil and iris. Although DeepVOG has the highest overall IoU score, one must also consider that the DeepVOG model is a two-class (binary) classifier (pupil vs. background) being compared against models of three-class segmentation (pupil, iris, background) and, in the former case, the IoU score is inflated by the presence of a large number or background pixels. This analysis also demonstrates that segmentation performance is improved by the inclusion of $\mathcal{L}_{COM}$ for all cases. Some examples of segmentation outputs with the inclusion of $\mathcal{L}_{COM}$ for OpenEDS and RIT-Eyes datasets are shown in Figure~\ref{fig:ground_truth_prection}.

\subsubsection{Qualitative Analysis: Effectiveness of $\mathcal{L}_{COM}$ loss}
\label{sec:activation_name}
\begin{figure*}
\begin{center}
\includegraphics[width=\linewidth]{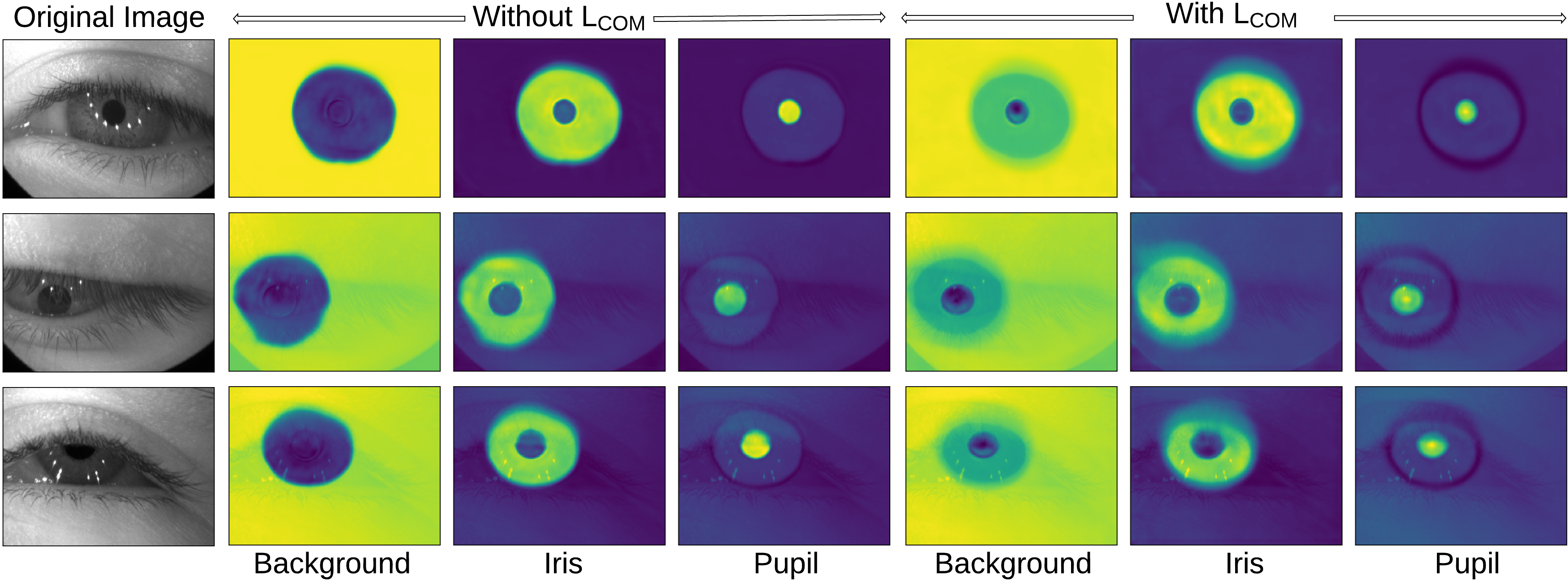}
\end{center}
\caption{Figure showing 2D activation maps. Columns (L-R): Original image (1st column), activation maps for background, iris and pupil class for model DenseElNet \textit{without} $\mathcal{L}_{COM}$ (2nd-4th column) \textit{with} $\mathcal{L}_{COM}$ (5th-7th column). Three rows show three different cases with bottom two having the original image in the background for reference. \textit{(Best viewed on screen)}}
\label{fig:activations}
\end{figure*}

\begin{figure}
\begin{center}
\includegraphics[width=\linewidth]{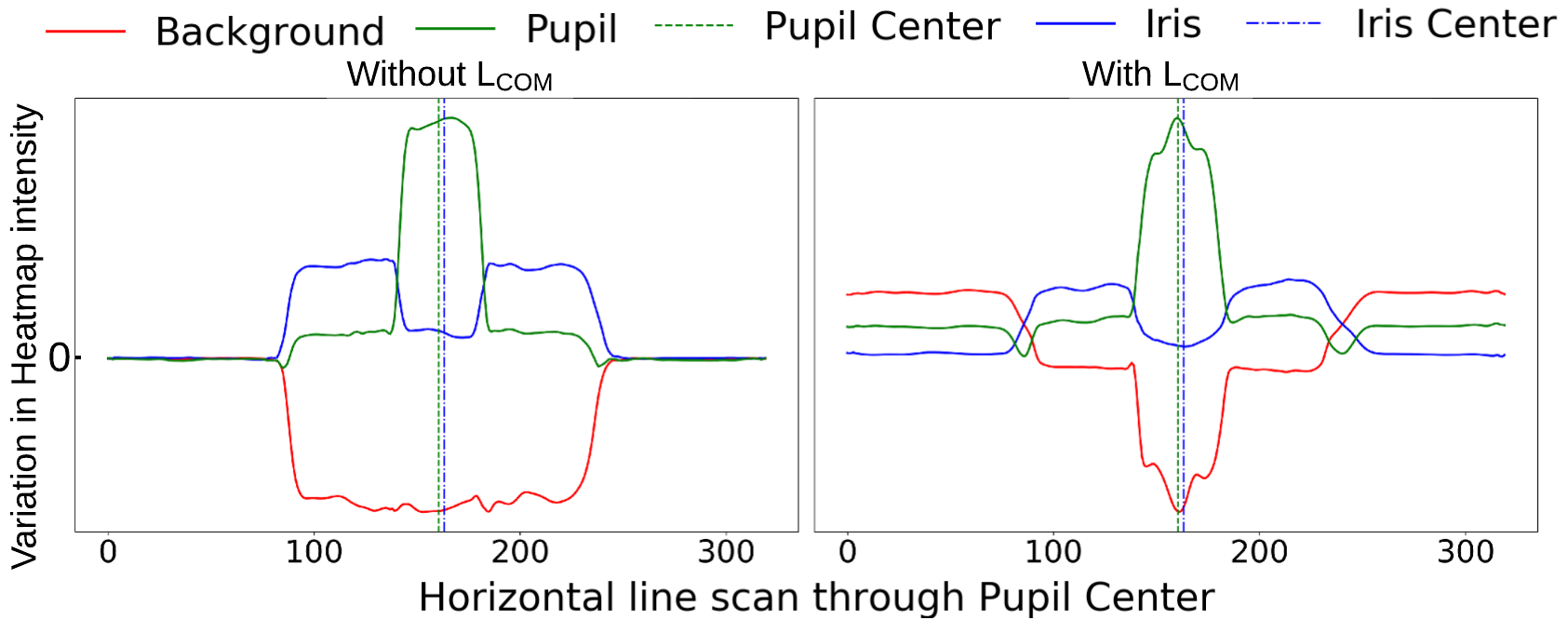}
\end{center}
\caption{A horizontal line scan across the pupil center to visualize DenseElNet output behavior without $\mathcal{L}_{COM}$ (left) and with $\mathcal{L}_{COM}$ (right). The inclusion of $\mathcal{L}_{COM}$ generates characteristic peaks which do not impede the task of semantic segmentation while effectively scaling output pixel activations near the predicted pupil and iris centers \textit{(Best viewed on screen)}.}
\label{fig:activations_line}
\end{figure}

Here, we study the impact of the $\mathcal{L}_{COM}$ loss function with the DenseElNet architecture. Figure~\ref{fig:activations} shows the activation maps generated (\textit{with} and \textit{without})  $\mathcal{L}_{COM}$ for three eye images. On closer observation of the pupil class, we observe a high intensity peak in the region around pupil center in the \textit{with} $\mathcal{L}_{COM}$ condition (last column) compared to the \textit{without} $\mathcal{L}_{COM}$ condition (fourth column from left). This peak around the pupil center is also evident in Figure~\ref{fig:activations_line} which shows a horizontal scan through the pupil center of one of the eye images illustrating the relative activation value for background, pupil, and iris \textit{without} (left) and \textit{with} (right) $\mathcal{L}_{COM}$.

Note that in Figure~\ref{fig:activations}, the iris activation maps appear even when the iris is occluded by the eyelids in both \textit{with} $\mathcal{L}_{COM}$ (second column from right) and \textit{without} $\mathcal{L}_{COM}$ (third column from left) conditions. 

Figure~\ref{fig:activations_line} shows relatively flat activation values near the iris centers for the iris class in both \textit{with} and \textit{without} $\mathcal{L}_{COM}$ cases; no peak is evident in the iris activation values. Note that the minimum in the background activation value localizes the center of the \textit{iris} representing the inverse of the background (non-iris) region.

%\subsection{Ablation Study}
\subsection{Center via bottleneck vs softargmax}
\begin{figure}
\begin{center}
\includegraphics[width=\linewidth]{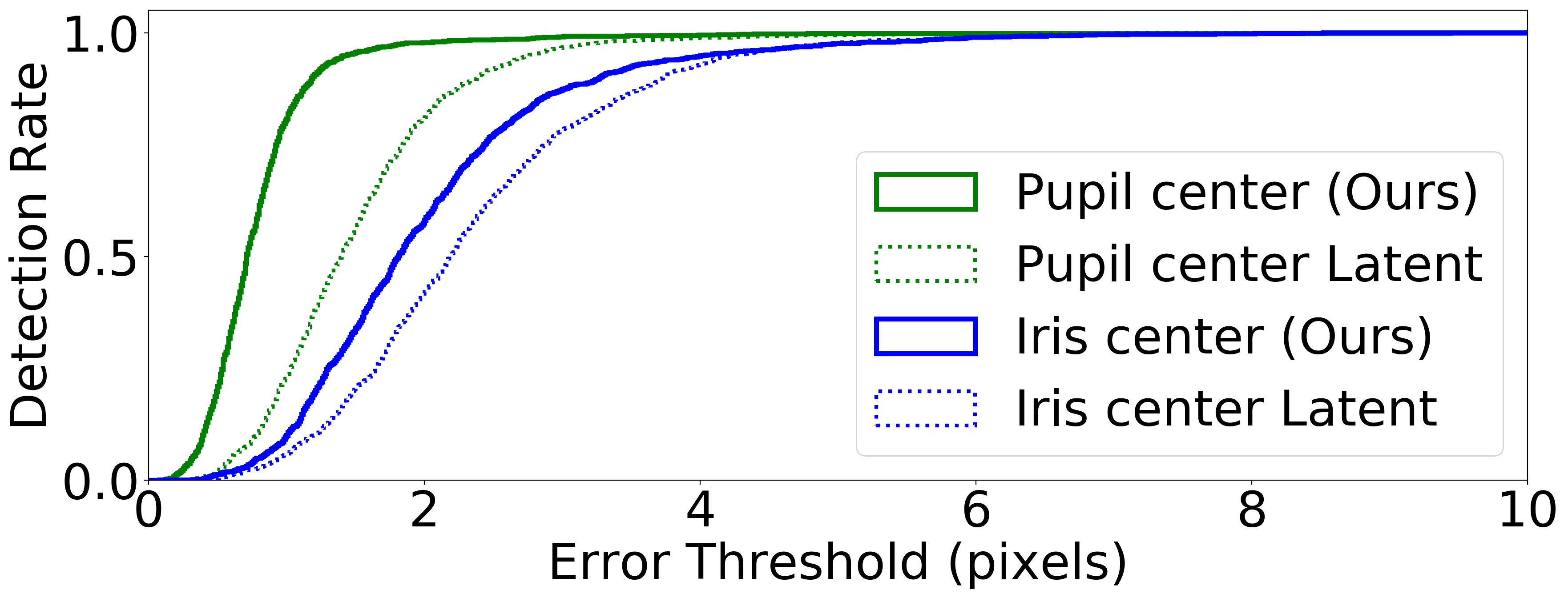}
\end{center}
\caption{The difference between pupil and iris detection rate in the OpenEDS dataset. Estimates are derived from the latent space and final segmentation maps (DenseElNet).}
\label{fig:compare_latent}
\end{figure}

%\subsection{Predicting centers via segmentation maps vs bottleneck regression}
\label{sec:center_bottlevssoftargmax}
To help provide an intuition regarding future network designs, we observe the impact of regressing the pupil and iris center estimates from the bottleneck (latent) layer~\cite{Fuhl2019a}, as opposed to estimating them using soft-argmax on the output segmentation maps (see Figure~\ref{fig:EllSeg_framework}). Estimates from segmentation outputs are observed to be better than those regressed from latent space (pupil 81\% $\rightarrow$ 98\% and iris 42\% $\rightarrow$ 58\% detection at the two-pixel error margin) (see Figure~\ref{fig:compare_latent}). We hope that this intuition can help guide future efforts for CNN based near-eye feature extraction.

\section{Summary}
%***Section has been revised***}
%\jeff{*** Why? The original Summary is better in some respects.  Instead of replacing it wholesale, I suggest *editing* the existing summary so that we can respond to the edits.}
%Our un-optimized implementation of DenseElNet-EllSeg and RITNet-EllSeg are benchmarked at 30Hz and 40Hz respectively on a NVIDIA 1080 Ti, Intel-7800K}.

This paper presents EllSeg, a new framework for training a CNN to directly segment the entire elliptical structures of the pupil and iris. This framework was applied to RITnet~\cite{chaudhary2019ritnet}, DeepVOG~\cite{Yiu2019b} and a custom designed hybrid model, DenseElNet, for segmentation as well as predicting pupil/iris ellipse estimates from eye images.

In Section~\ref{sec:state_of_art_models}, we benchmark our custom designed network architecture, DenseElNet, and achieve better baseline PartSeg performance to state-of-the-art encoder-decoder architectures, RITnet and DeepVOG (see Table~\ref{tab:compare_network_archs}). Our un-optimized forward pass implementation of DenseElNet operates at atleast 120Hz on a NVIDIA 1080 Ti, Intel-7800K. In Section~\ref{sec:center_estimate}, we show that our proposed framework \textit{EllSeg} outperforms part-segmentation networks, \ie \textit{PartSeg}, for pupil \rev{(OpenEDS: 0.2$\%$, NVGaze: 11$\%$, RIT-Eyes: 12$\%$)} and iris center \rev{(OpenEDS: 4$\%$, NVGaze: 29$\%$, RIT-Eyes: 25$\%$)} detection across three test datasets. Additional analysis reveals that the accuracy of EllSeg can be attributed to greater robustness to occlusion of the iris and pupil by the eyelids. 

Section~\ref{sec:improve_ellipse_estimate} demonstrates that the addition of $\mathcal{L}_{COM}$ loss function to the EllSeg framework results in improved pupil/iris ellipse estimates for pupil (\rev{OpenEDS: 2$\%$, NVGaze: 11$\%$, RIT-Eyes: 21$\%$}) and iris center (\rev{OpenEDS: 15$\%$, NVGaze: 29$\%$, RIT-Eyes: 40$\%$}) detection rate within a two-pixel error margin) and segmentation performance ($>0.6\%, >1.5\%, >2\%$ for OpenEDS, NVGaze and RIT-Eyes respectively).

Visual inspection of output EllSeg activation maps reveals high confidence conditioned around the pupil and iris centers. Lastly in Section~\ref{sec:center_bottlevssoftargmax}, we determine that deriving pupil and iris centers using softargmax is better than regressing the same via the bottleneck layer.

\section{Conclusion and future work}
%***Section has been revised***}

To conclude, we present EllSeg, a simple 3-class full ellipse segmentation framework intended to extend conventional encoder-decoder architectures for the segmentation of eye images into pixels that represent the pupil, iris, and background. The EllSeg framework was benchmarked on multiple datasets using two network architectures: RITnet and our custom CNN design, DenseElNet. Results demonstrate superior estimation of the pupil and iris centers and orientation compared to their eye part segmentation models. An added benefit of the EllSeg framework is that it extends model training to image datasets in which only the pupil center has been labelled. Superior performance by the EllSeg framework can be attributed to greater robustness to occlusion of the pupil or iris. 

While we evaluate EllSeg on multiple datasets collected from a large pool of individuals (see Table~\ref{tbl:existing_datasets}), a user based evaluation was not performed due to the time consuming nature of manual data collection and labelling. For future work, we intend on performing a comprehensive user study of our model on a wide range of subjects to further quantify the performance of our framework. We also intend on exploring other models with varying complexity to evaluate the efficacy of EllSeg. Pretrained models, code and other related resources will be made publicly available~\footnote{\url{https://cis.rit.edu/~rsk3900/EllSeg/}}.

% In this work, we present EllSeg, a simple 3-class full ellipse segmentation framework, and demonstrate superior ellipse shape and center estimation as compared to the standard 4-class eye parts segmentation. We benchmark the EllSeg framework on multiple datasets using two network architectures, RITnet and our custom CNN design; DenseElNet. The EllSeg framework enables us to derive ellipse-center estimates which fall within a median score of 0.8 pixels (for image size of 320x240) from the groundtruth using segmentation output maps. The EllSeg framework combats occlusion due to eyelids along with general improvements observed across multiple metrics. Furthermore, this formulation enables joint learning on multiple eye image datasets with only the pupil center labelled. Pretrained models will be made publicly available.
%~\footnote{\url{http://www.cis.rit.edu/~rsk3900/EllSeg/}}. 

\section{Acknowledgements}
We thank Research Computing~\cite{https://doi.org/10.34788/0s3g-qd15} at the Rochester Institute of Technology for providing all necessary hardware required for this project. We also thank Dr. Christopher Kanan for his helpful feedback and guidance. Lastly, we would like to thank Dr. Thiago Santini and Dr. Wolfgang Fuhl for their guidance in setting up the framework for their PuRe and PuReST algorithms.
\FloatBarrier

\bibliographystyle{unsrt}
\bibliography{main}

\newpage
\FloatBarrier

\end{document}